\documentclass[conference]{IEEEtran}
\IEEEoverridecommandlockouts

\usepackage{cite}
\usepackage{amsmath,amssymb,amsfonts}
\usepackage{algorithmic}
\usepackage{graphicx}
\usepackage{textcomp}
\usepackage{xcolor}

\usepackage{subfigure}

\usepackage{booktabs}
\usepackage{multirow}
\usepackage{enumerate}
\usepackage{color,xcolor}
\usepackage{caption}

\def\BibTeX{{\rm B\kern-.05em{\sc i\kern-.025em b}\kern-.08em
    T\kern-.1667em\lower.7ex\hbox{E}\kern-.125emX}}
\begin{document}

\title{Temporal Dynamics Decoupling with Inverse Processing for Enhancing Human Motion Prediction
}


\author{\IEEEauthorblockN{Jiexin Wang, Yiju Guo, Bing Su*\thanks{* Corresponding author.}}
\IEEEauthorblockA{\textit{Gaoling School of Artificial Intelligence, Renmin University of China, Beijing, China}}
}

\maketitle

\begin{abstract}
Exploring the bridge between historical and future motion behaviors remains a central challenge in human motion prediction. While most existing methods incorporate a reconstruction task as an auxiliary task into the decoder, thereby improving the modeling of spatio-temporal dependencies, they overlook the potential conflicts between reconstruction and prediction tasks.
In this paper, we propose a novel approach: Temporal Decoupling Decoding with Inverse Processing (\textbf{$TD^2IP$}). Our method strategically separates reconstruction and prediction decoding processes, employing distinct decoders to decode the shared motion features into historical or future sequences. Additionally, inverse processing reverses motion information in the temporal dimension and reintroduces it into the model, leveraging the bidirectional temporal correlation of human motion behaviors. 
By alleviating the conflicts between reconstruction and prediction tasks and enhancing the association of historical and future information, \textbf{$TD^2IP$} fosters a deeper understanding of motion patterns.
Extensive experiments demonstrate the adaptability of our method within existing methods.
\end{abstract}


\section{Introduction}
\label{sec:intro}

In recent years, we have witnessed a significant increase in research efforts that apply deep learning to Human Motion Prediction (HMP), which is a computer vision task with many potential applications~\cite{koppula2013anticipating,unhelkar2018human,paden2016survey,rasouli2019pie,chen20203d,tang2020uncertainty}. HMP aims to predict the future motion sequence of the skeleton-based human body based on a given historical motion sequence~\cite{ryoo2011human,kong2022human}.

\begin{figure}[!t]
\begin{center}
\subfigure[Prediction task only.]{
\includegraphics[width=0.33\columnwidth]{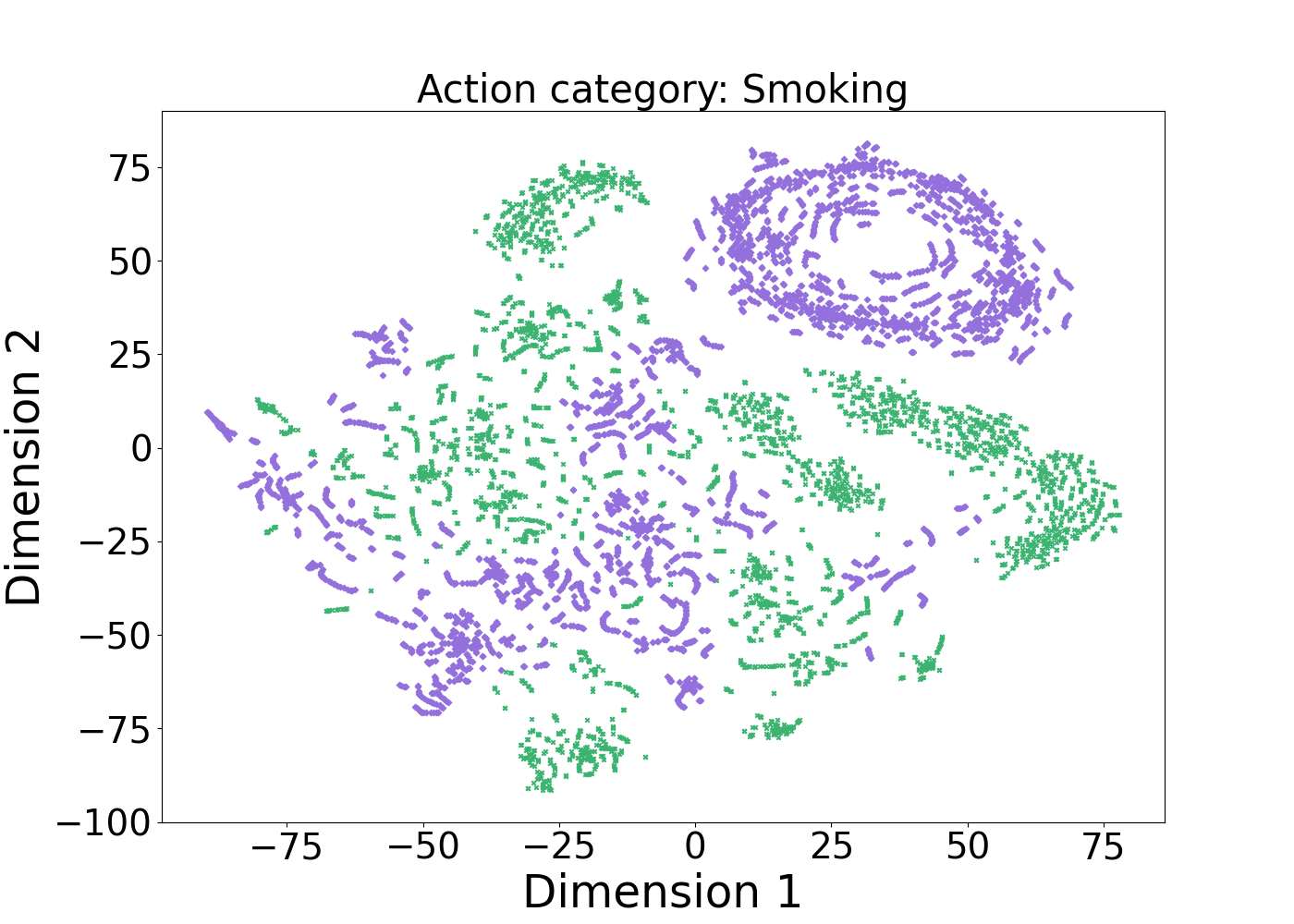}
\includegraphics[width=0.33\columnwidth]{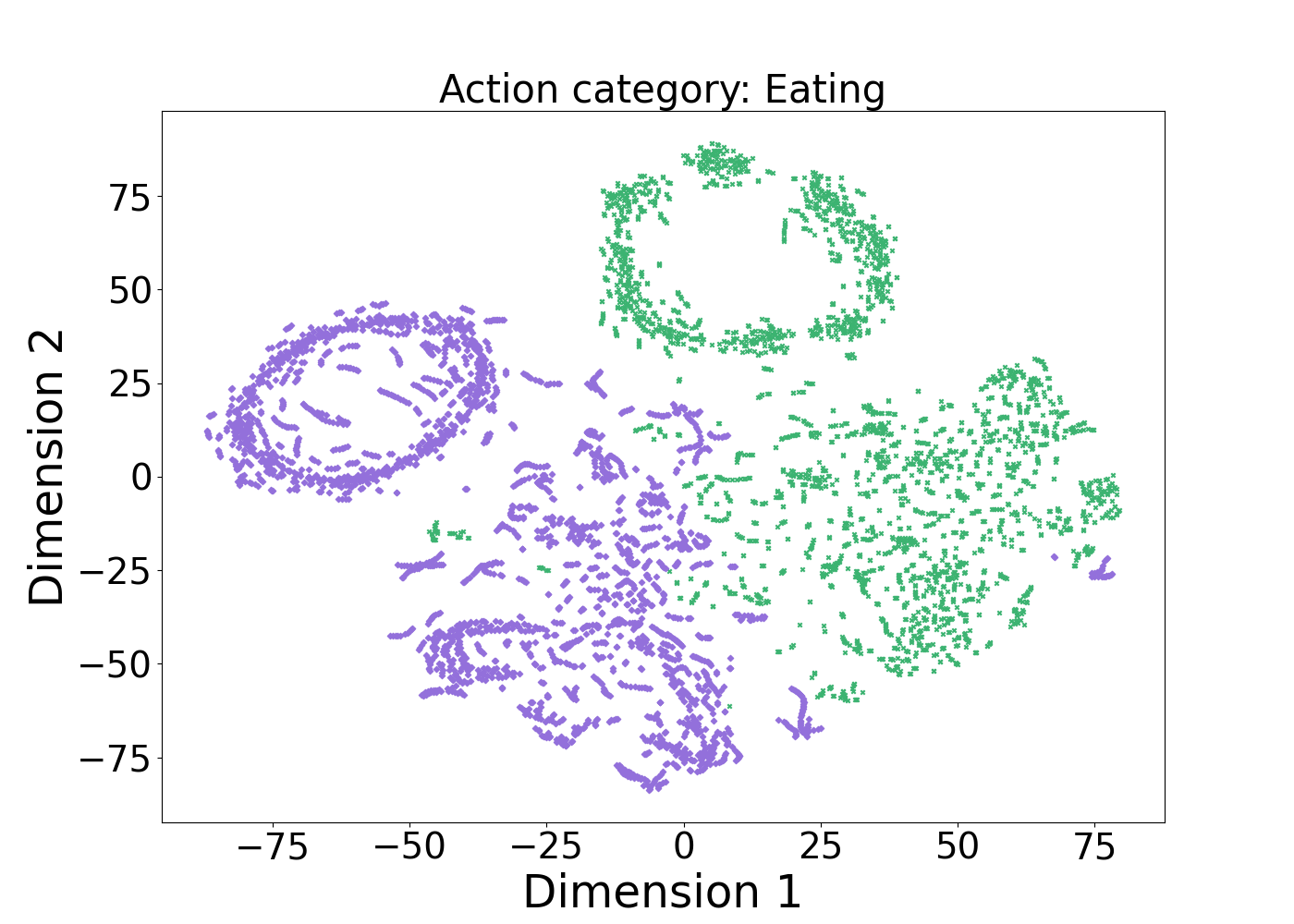}
\includegraphics[width=0.33\columnwidth]{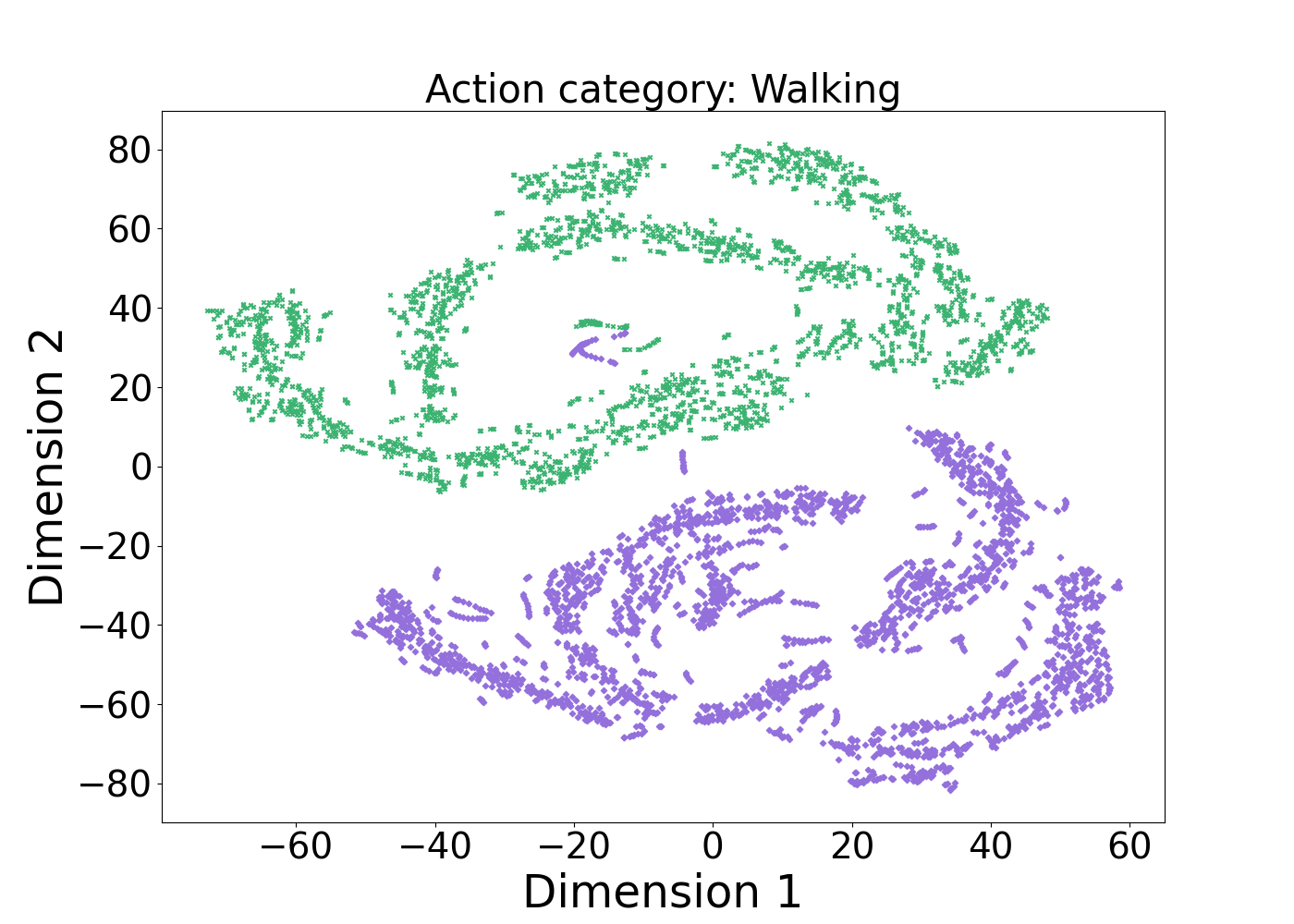}
}
\subfigure[Prediction and reconstruction tasks.]{
\includegraphics[width=0.33\columnwidth]{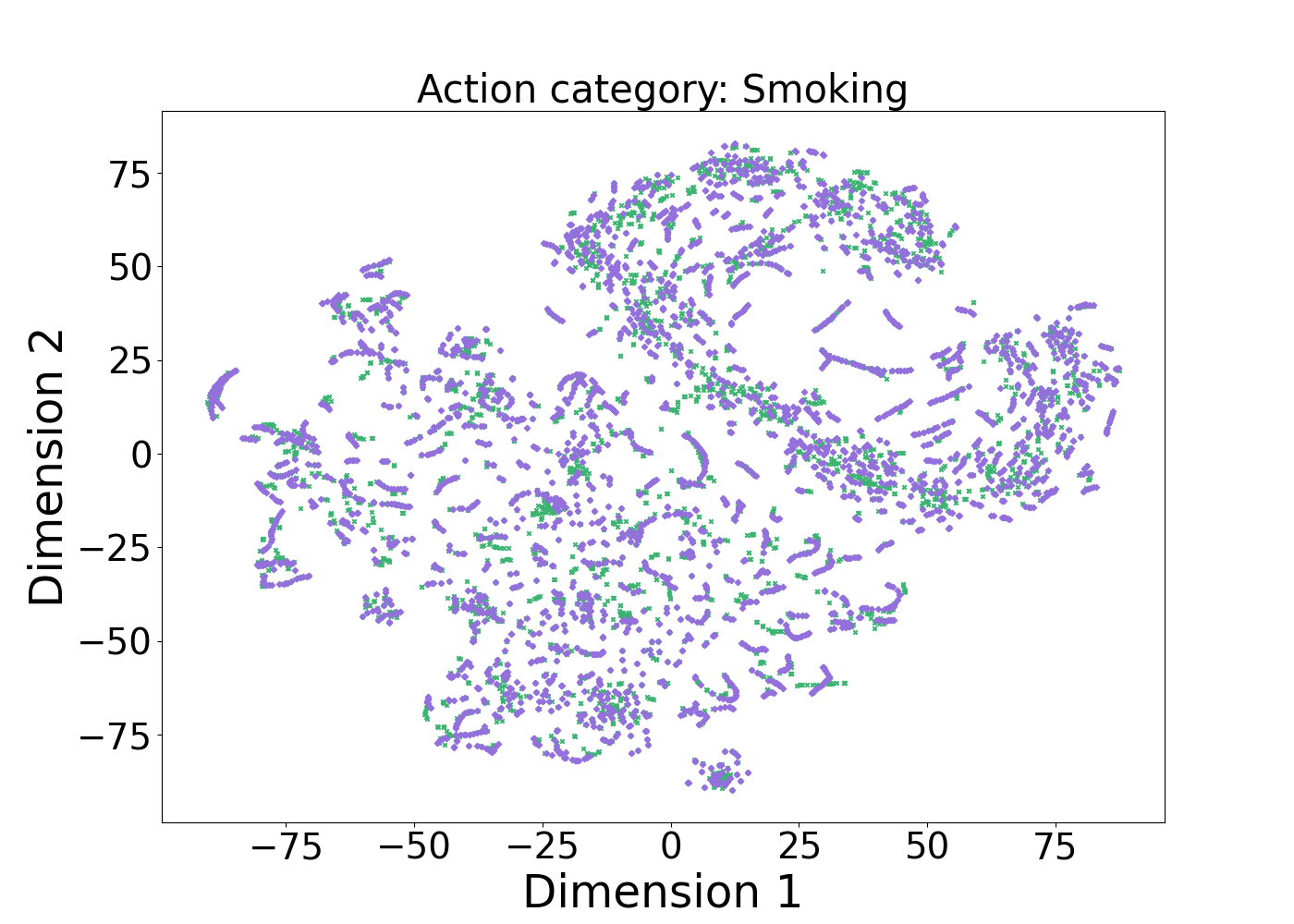}
\includegraphics[width=0.33\columnwidth]{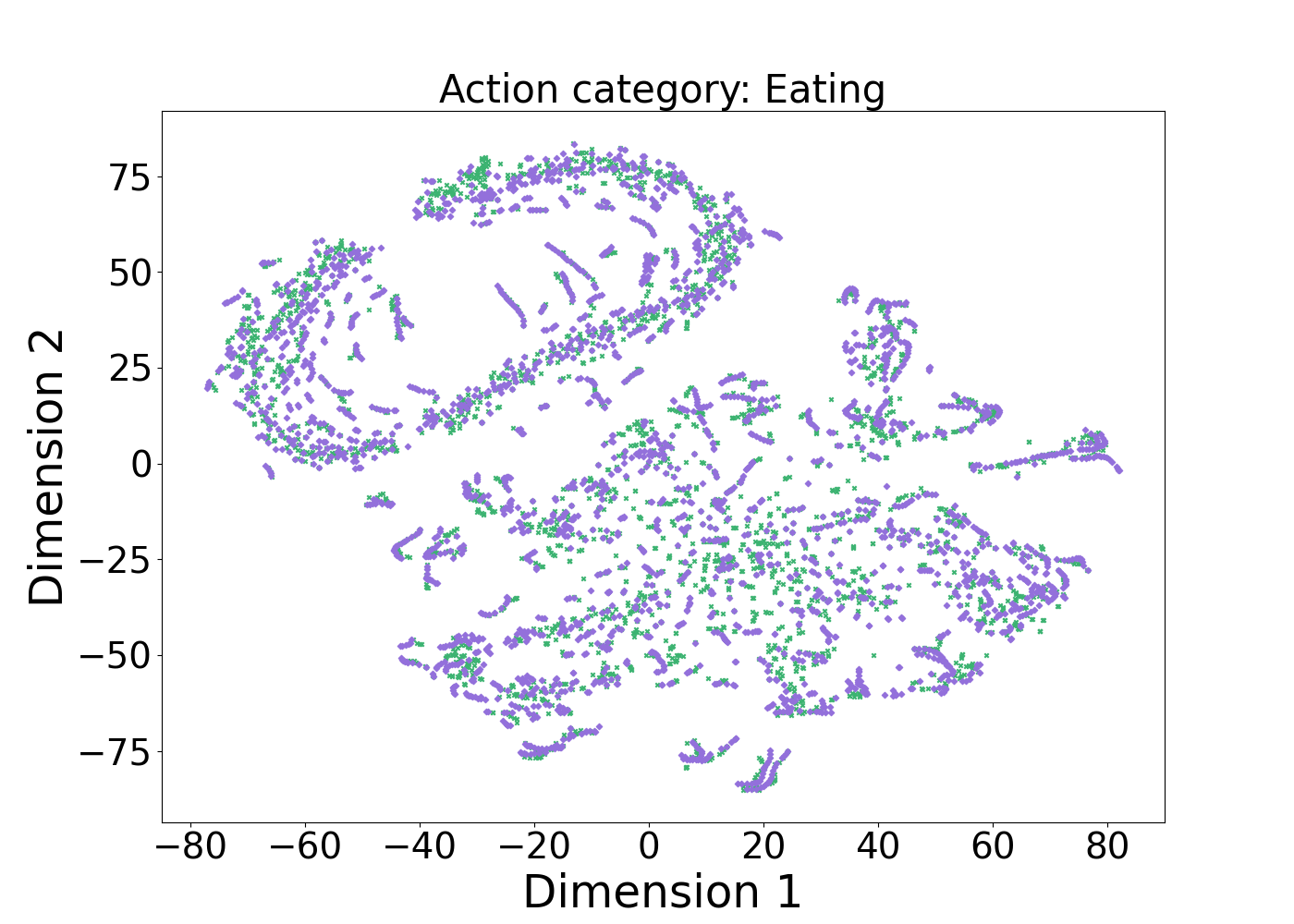}
\includegraphics[width=0.33\columnwidth]{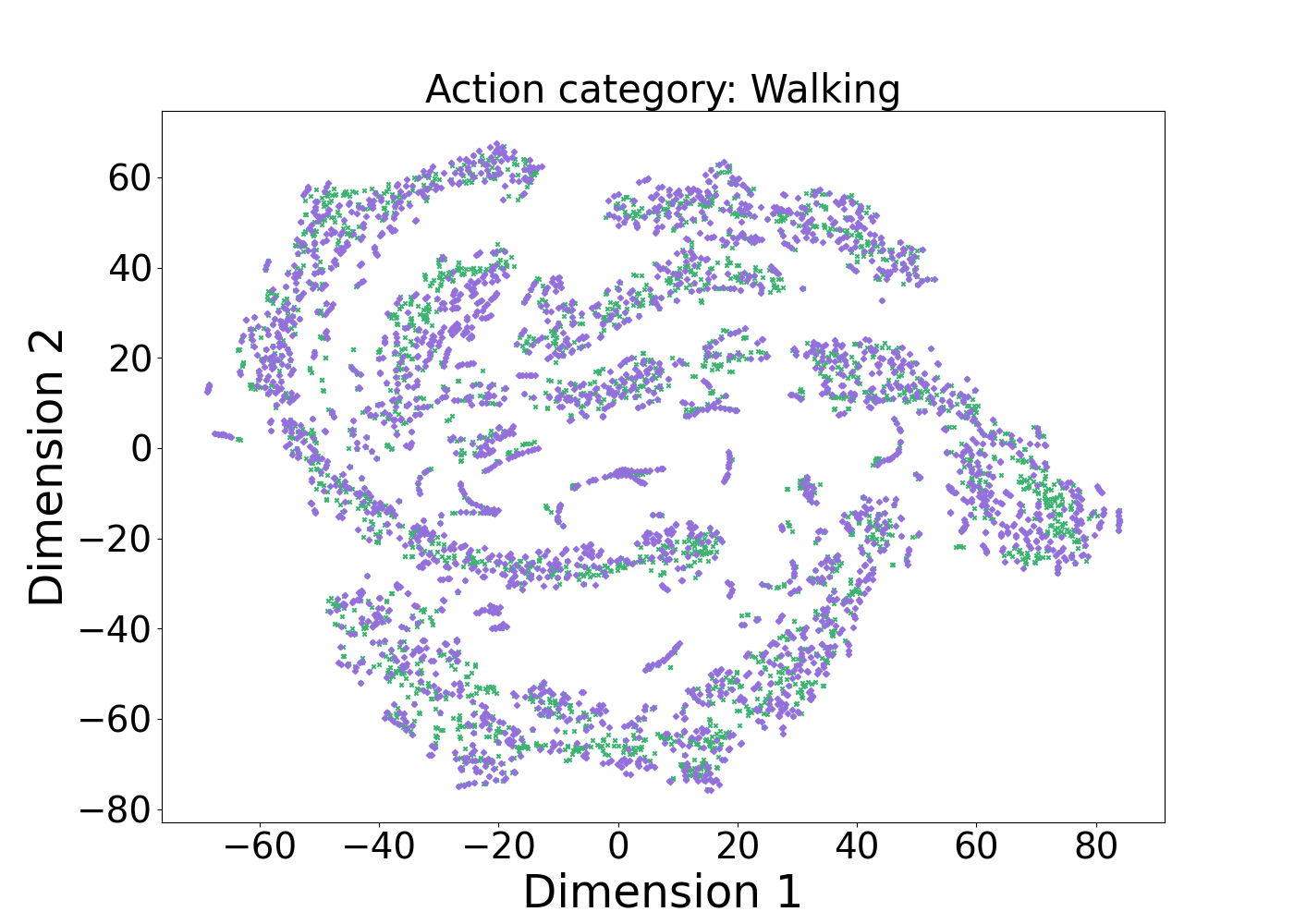}
}
\vspace{-0.1in}
\caption{T-SNE visualization of motion features in H3.6M under different networks (Left: GCN, Middle: LSTM, Right: Transformer). Purple points denote the ground truth motion features, while green points indicate the predicted features. Incorporating the reconstruction task effectively enhances the alignment between predicted and ground truth motion features.
}
\label{fig:mot_0}
\end{center}
\vspace{-0.20in}
\end{figure}

\begin{figure}[!t]
\begin{center}
\vspace{-0.1in}
\subfigure[Reconstruction performance.]{\label{fig:mot_1}
\includegraphics[width=0.33\columnwidth]{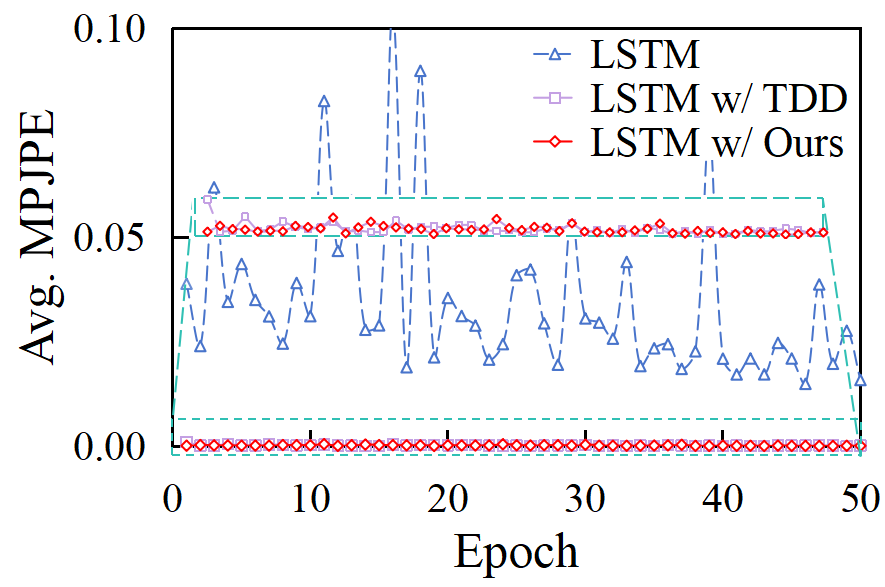}
\includegraphics[width=0.33\columnwidth]{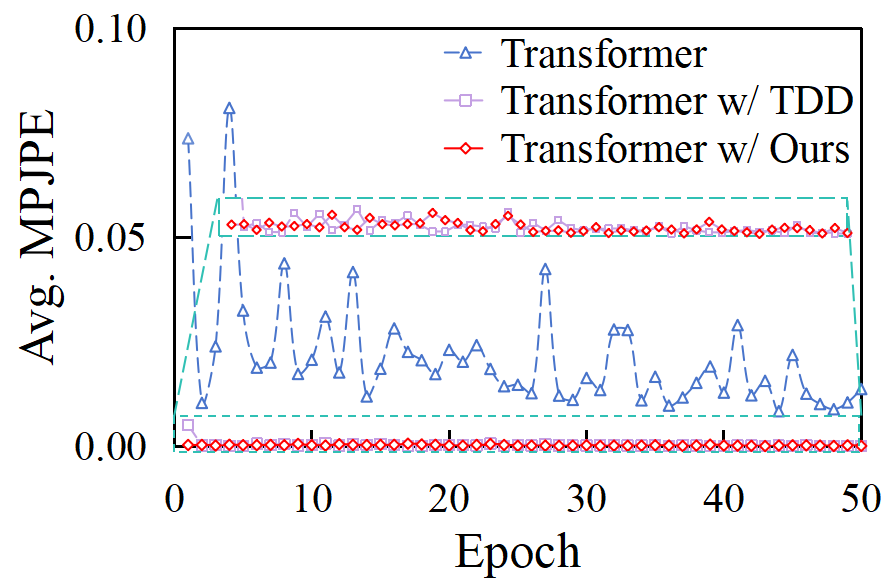}
\includegraphics[width=0.33\columnwidth]{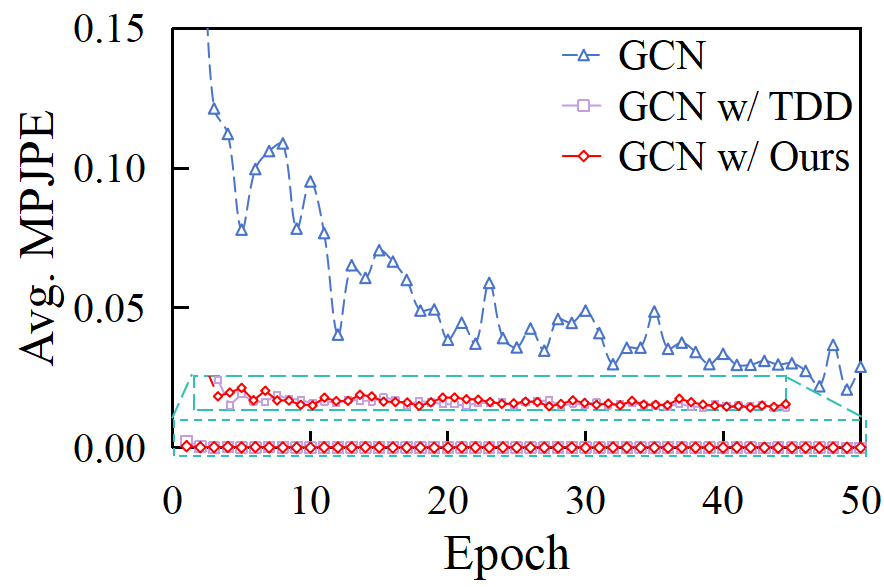}
}
\vspace{-0.1in}
\subfigure[Prediction performance.]{\label{fig:mot_2}
\includegraphics[width=0.33\columnwidth]{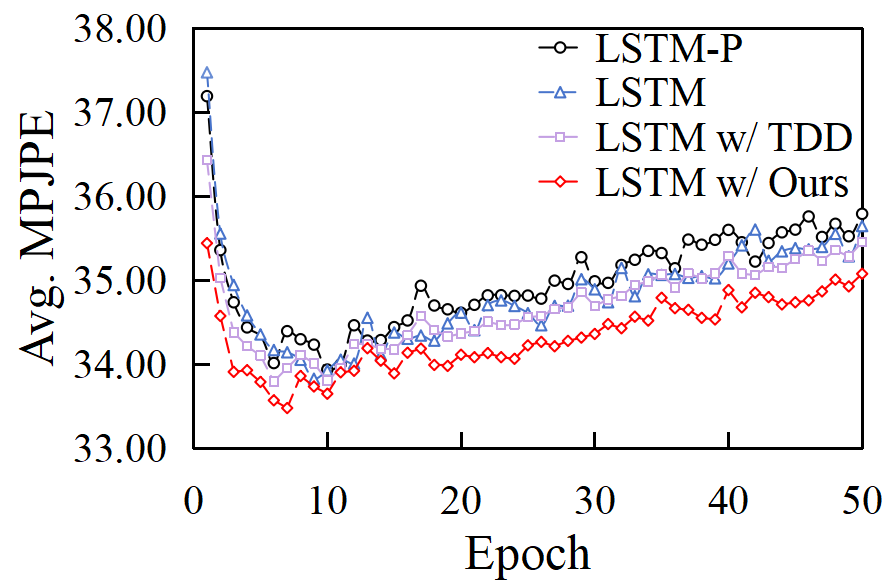}
\includegraphics[width=0.33\columnwidth]{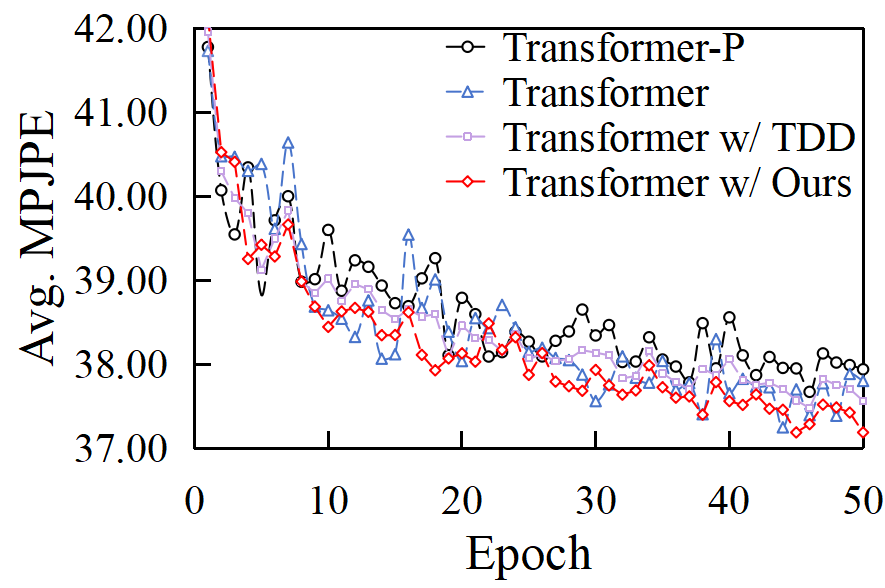}
\includegraphics[width=0.33\columnwidth]{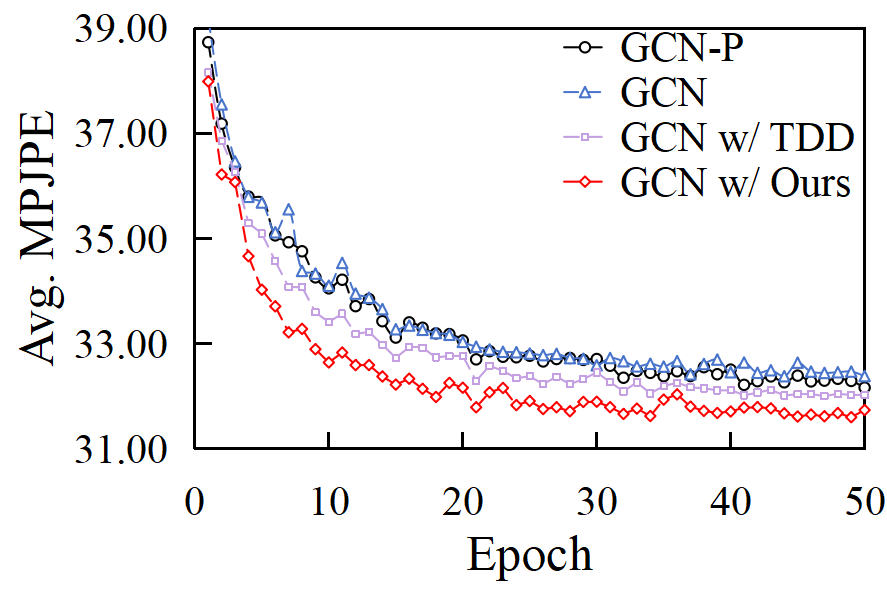}
}
\vspace{-0.05in}
\caption{Comparison of predictive performance (test loss) in H3.6M under different networks. ``LSTM'', ``Transformer'', and ``GCN'' simultaneously perform both reconstruction and prediction tasks with a shared decoder. ``-P'' indicates models solely performing the prediction task.
}
   \label{fig:mot}
\end{center}
\vspace{-0.2in}
\end{figure}

Early studies in HMP focused on extracting motion representations from historical sequences to generate predictions directly~\cite{pavlovic2000learning,wang2005gaussian,taylor2006modeling,lehrmann2014efficient,su2016hierarchical}.
Most existing methods enhance the prediction performance by incorporating a reconstruction task as an auxiliary task into the decoder~\cite{mao2019learning,mao2020history,li2021skeleton,coll2022representing,li2022skeleton,xu2023auxiliary}. This auxiliary task is closely related to the primary prediction task, as both require the network to model spatial-temporal dependencies from different perspectives effectively. 
Besides, coupling both tasks into a shared feature aims to enhance the correlation between historical and future information, requiring a more expressive motion representation. 
Therefore, the additional requirements imposed by the auxiliary task force the network to learn more effective and comprehensive motion representations, as shown in Fig.\ref{fig:mot_0}.
However, the impact of the auxiliary task on predictive performance varies across different networks, as shown in Fig.\ref{fig:mot_2}.
Existing methods have focused primarily on network structure and feature extraction, overlooking the potential interference and conflicts between the reconstruction and prediction tasks, which have yet to be fully explored.

According to the analysis in stacked denoising autoencoders~\cite{vincent2010stacked} and SimMTM~\cite{dong2024simmtm}, the reconstruction task involves projecting the motion features extracted from the historical behaviors back onto the manifold of the original historical behaviors. In contrast, the prediction task projects these features onto the manifold of future behaviors. Directly incorporating reconstruction into the decoder can bridge past and future motion behaviors more effectively and enhance the correlation between the decoder and motion data. However, this dual-task approach may fail as the decoder must balance between two manifolds, potentially leading to insufficient feature expression for certain time horizons. 
Moreover, the inherent imbalance in task difficulty, as shown in Fig.\ref{fig:mot}, highlights the inefficiency of allocating equal attention to both tasks. Handling multiple tasks also concurrently imposes an additional learning burden on the decoder. This raises a natural question: can we further improve the predictive performance by comprehensively considering the above aspects?

Based on the above motivations, we go beyond the straightforward coupling of the reconstruction and prediction tasks within a shared decoder and propose a natural idea called Temporal Decoupling Decoding ($TDD$),  serving as a complementary augmentation to existing methods. Technically, we propose a decoding architecture explicitly separating the reconstruction and prediction processes, with each decoding component specialized for either reconstructing historical motion or predicting future motion. This segregation mitigates the risk of the model showing bias towards one task over the other, enhancing the expression of motion feature, as shown in Fig.~\ref{fig:mot}.
Furthermore, this methodology is crucial in capturing the intricate dynamics across different temporal horizons, thereby cultivating a deeper comprehension of the fundamental motion patterns.

Additionally, the core idea of HMP revolves around comprehending the interplay between historical and future motion behaviors, aiming to seek an efficient bridging through learned mapping functions. 
Our decoupled decoding architecture, where each decoder targets a specific temporal horizon, presents a challenge in maintaining global temporal correlations between historical and future behaviors.
Recognizing the importance of bidirectional correlation, where future motion should remain coherent with historical observations even when sequences are reversed, we propose a novel auxiliary task called Inverse Processing ($IP$) inspired by reverse engineering principles~\cite{baxter1997reverse,d2000genetic,raja2007reverse,dijkstra2021predictive}. $IP$ entails incorporating a unique auxiliary task into the training scheme of our method. Specifically, after predicting future motion based on historical input, we reverse the future motion in time and reintroduce it into the model, enabling the prediction of historical motion. 
$IP$ enhances the correlation between historical and future information in both temporal directions and allows each decoder of our method to access the complete motion information, fostering a comprehensive understanding of motion behavior. 
Through extensive experiments and comparisons with state-of-the-art approaches, we demonstrate the effectiveness and superiority of our method. 
The main contributions of this paper are summarized below:
\begin{itemize}
\item We propose a novel extension to the mainstream Encoder-Decoder framework that employs a shared decoder for generating both historical and future behaviors. Our approach decouples the decoding process for reconstruction and prediction. Each decoder specializes in a specific time horizon of historical or future information, effectively alleviating the interference and conflicts of different tasks.
\item We propose a novel auxiliary task termed inverse processing, which enables the model to leverage future motion information to predict historical information. This mechanism significantly enhances the correlation between historical and future information, leading to a more comprehensive understanding of human motion behavior.
\item  We conduct extensive experiments by integrating our method with existing approaches on standard HMP benchmarks. Experiments validate that our method can effectively improve the prediction performance.

\end{itemize}

\section{Methodology}
\label{sec: Methodology}

\subsection{Problem Formulation}
Given $\mathbf{X}=[X_1,\cdots,X_{T_p}] \in \mathbb{R}^{T_p \times J \times 3}$, where $X_t \in \mathbb{R}^{J \times 3} $ represents the 3D coordinates of $J$ body joints in time $t$, and the target future pose sequence $\mathbf{Y}=[X_{T_p+1},\cdots,X_{T_p+T_f}] \in \mathbb{R}^{T_f \times J \times 3}$. $T = T_p + T_f$ is the total length of motion. Formally, the HMP problem aims to predict $\mathbf{Y}$ given $\mathbf{X}$. The primary challenge in HMP is to devise an effective predictor $\mathcal{F}_\mathrm{{pred}}(\cdot)$ such that the predicted future motion $\hat{\mathbf{Y}}=\mathcal{F}_\mathrm{{pred}}(\mathbf{X})$ is as close to the ground-truth future motion $\mathbf{Y}$ as possible.
\begin{figure}[!t]
    \centering
    \includegraphics[width=0.99\linewidth]{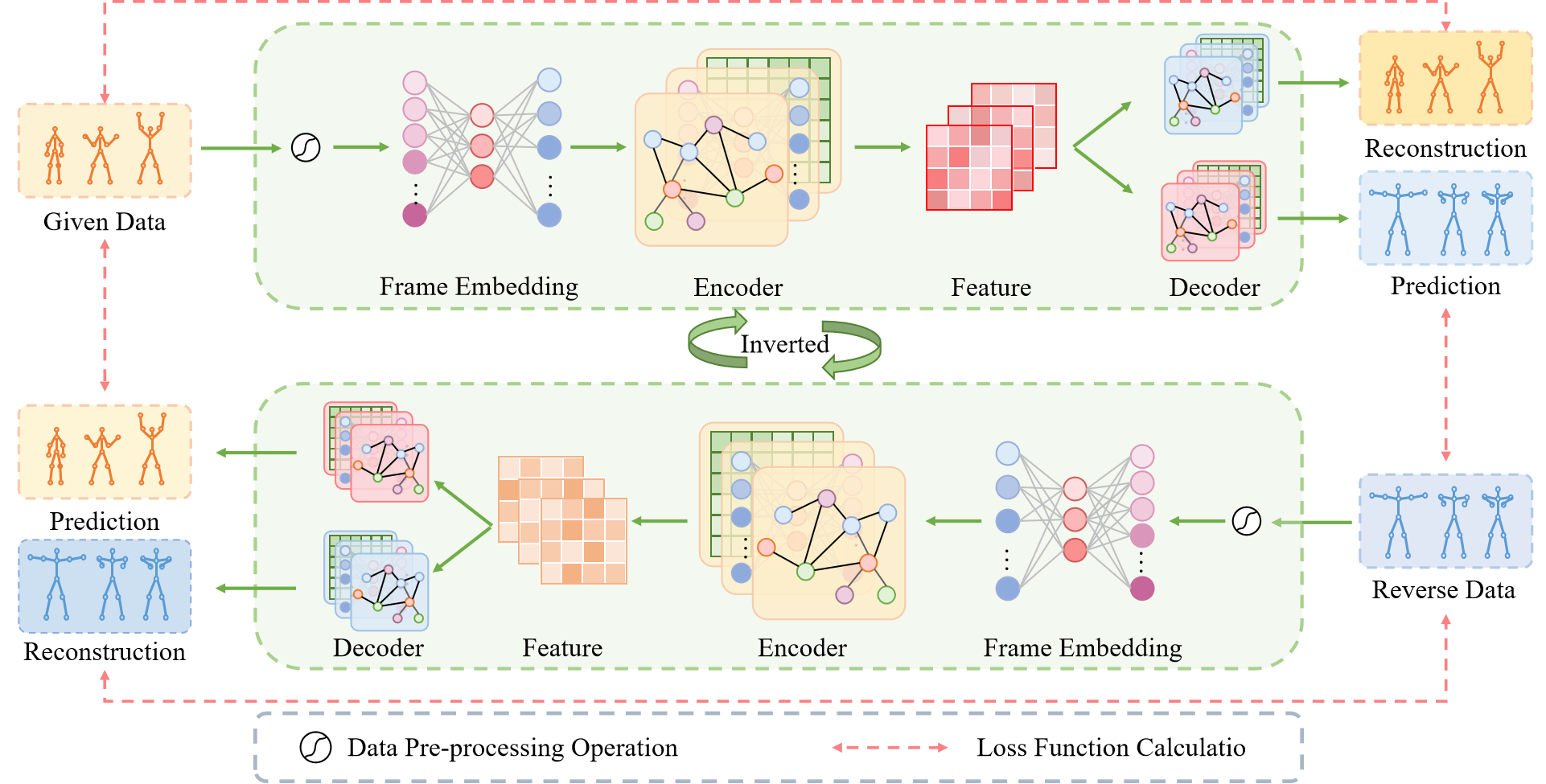}
    \caption{Illustration of the $TD^2IP$. 
    }
    \label{fig:model}
    \vspace{-0.20in}
\end{figure}

\subsection{Framework Architecture}
In this work, we propose an innovative learning framework named $TD^2IP$ for HMP, as shown in~\figurename{~\ref{fig:model}}. The core idea of the framework is to enhance the prediction models by decoupling the reconstruction and prediction processes, coupled with improving the correlation between historical and future information through inverse processing. Specifically, for a given input sequence $\mathbf{X}$, a prediction module $\mathcal{F}_\mathrm{{pred}}$ consists of a embedding to project $\mathbf{X}$ into a feature space $\hat{X} = W_{2} ( \sigma ( W_{1}  \mathbf{X} + b_1) ) + b_2$, an encoder $\varphi$ for modeling spatio-temporal dependencies in the motion data $M = \varphi(\hat{X})$, and a decoder $g$ for generating the prediction $\hat{\mathbf{Y}} = g(M)$. During decoding, we leverage Temporal Decoupling Decoding ($TDD$) to project the shared motion representation into historical or future time horizons, respectively, effectively mitigating the interference and conflicts between different tasks. Furthermore, drawing inspiration from the bidirectional correlation of human motion behavior and reverse engineering principles, we propose Inverse Processing ($IP$) to enhance the model's capacity to bridge historical and future information.

\noindent\textbf{Temporal Decoupling Decoding.}
In the mainstream Encoder-Decoder frameworks of HMP, a common practice is to employ a single decoder to reconstruct historical motion and predict future motion simultaneously. This shared decoder strategy aims to enhance the correlation between historical and future information, encouraging the model to learn more comprehensive representations. 
However, the simultaneous execution of both tasks within one decoder may introduce interference and conflicts, potentially limiting the auxiliary benefits of reconstruction and even detracting from the prediction task.

To tackle this challenge, we decompose the decoding process within the Encoder-Decoder framework into two distinct components: reconstruction decoding and prediction decoding. This separation allows different decoding processes to focus on specific time horizons, effectively isolating the primary prediction task from the auxiliary reconstruction task. Consequently, this unleashes the potential of the decoder and learned motion features. Specifically, we propose Temporal Decoupling Decoding ($TDD$), consisting of two decoders, $(g_h, g_f)$, each assigned to decode the shared representations $M$ into the historical of future time horizons. For each decoder $g_k$, where $k=\{h,f\}$, the transformation of $M$ yields an output denoted as $P_k$.
Both decoders share the same structure, differing only in the temporal dimension of the output. The decoding process is as follows:

\begin{equation}
P_k = g_k(M) \in \mathbb{R}^{T_k \times J \times D},
\label{decoder}
\end{equation}
where $T_k$ denotes the reconstruction or prediction horizon specific to $g_k$. Each decoder follows the general decoding framework specified in Eq.~\ref{decoder}, generating sequences for different time horizons. With the deployment of the reconstruction and prediction decoders in $M$, $TDD$ generates reconstruction sequences $P_h \in \mathbb{R}^{T_p \times J \times D}$ and prediction sequences $P_f \in \mathbb{R}^{T_f \times J \times D}$.
The final prediction is obtained as:
\begin{equation}
\hat{\mathbf{Y}}_f = [P_h, P_f] \in \mathbb{R}^{T \times J \times D}.
\end{equation}
$[,]$ indicates the concatenation of vectors in the time dimension, with the ground truth motion denoted as $\mathbf{Y}_f=[\mathbf{X},\mathbf{Y}]$.

\noindent\textbf{Inverse Processing.} 
While $TDD$ alleviates the conflicts between the reconstruction and prediction tasks, each decoder focuses on a specific temporal horizon, which may lead to an incomplete understanding of human behaviors and weaken global temporal correlations between historical and future behaviors.
Historical motion forms the basis for future behaviors, and future information can also serve as prior knowledge for inferring historical data. This bidirectional temporal correlation underscores the importance of considering information linkage in both directions. 
Moreover, in reverse engineering, gaining insights into the target's evolution through reverse inference is crucial for enhancing the cognitive understanding of the system or expert. Discerning the dynamics and changes in the target's evolution proves instrumental in refining and improving the quality of the learned mapping function. 

Motivated by the above analysis, we propose Inverse Processing ($IP$) during the training to enhance the model's comprehension of human behaviors. The comprehensive training scheme involving $IP$ is depicted in~\figurename{~\ref{fig:model}}. Specifically, building upon standard prediction, we integrate an inverse prediction process that reverses the motion information backward in time. This enables the model to acquire bidirectional associations, capturing the temporal dynamics of human motion from future-to-past perspectives. The integration of $IP$ contributes to a more robust and nuanced understanding of motion behaviors, elevating the model's predictive capabilities.

During training, the motion data is denoted as $\mathbf{P}=[\mathbf{X},\mathbf{Y}]\in \mathbb{R}^{T \times J \times 3}$, and we perform a temporal flipping operation to obtain $\mathbf{P}_{r}=[X_{T},X_{T-1},\cdots,X_1]\in \mathbb{R}^{T \times J \times 3}$. By redividing the data, we obtain $\mathbf{X}_r=[X_T,\cdots,X_{T-T_p+1}] \in \mathbb{R}^{T_p \times J \times 3}$.
Similar to the standard prediction process, based on $\mathbf{X}_r$, the model's output is obtained as follows:

\begin{equation}
\hat{\mathbf{Y}}_r = [P_{h,r}, P_{f,r}] \in \mathbb{R}^{T \times J \times D}.
\end{equation}
Here, $P_{h,r}=g_h (M_r) $, $P_{f,r}=g_f (M_r) $, and $M_r= \varphi (W_{2} \left( \sigma \left( W_{1} \mathbf{X}_r + b_1) \right) + b_2\right)$ represents the encoded representation of $\mathbf{X}_r$. The ground truth motion denoted as $\mathbf{Y}_r=[X_T,\cdots,X_1]$.
This design enables each decoder to access the complete motion information and the model to learn from temporally reversed sequences, leading to a deeper understanding of bidirectional temporal correlations in human motion.

\subsection{Loss Function}
Consider a predictive model with an initial loss function $\mathcal{L}_1$, which can represent the Mean Per Joint Position Error (MPJPE) or incorporate additional regularization terms. For analytical purposes, we focus on the scenario where MPJPE serves as the original loss function. Formally, for a single training sample, the loss is expressed as follows:
\begin{equation}
    \mathcal{L}_f = \frac{1}{T\cdot J} \sum_{t=1}^{T} \sum_{j=1}^{J} \parallel \hat{\mathbf{Y}}_{f,t,j} - \mathbf{Y}_{f,t,j} \parallel^2,
\end{equation}
\begin{equation}
    \mathcal{L}_r = \frac{1}{T\cdot J} \sum_{t=1}^{T} \sum_{j=1}^{J} \parallel \hat{\mathbf{Y}}_{r,t,j} - \mathbf{Y}_{r,t,j} \parallel^2,
\end{equation}
where $\mathcal{L}_f$ represent the loss in the forward prediction, and $\mathcal{L}_r$ represents the loss in the reversed prediction.
To train the model, we combine $\mathcal{L}_f$ and $\mathcal{L}_r$ into a final loss function $\mathcal{L}$:
\begin{equation}
    \mathcal{L} = \mathcal{L}_f +  \mathcal{L}_r.
\end{equation}

\section{Experiments}
\label{sec:experiment}

\begin{table}[!t]\scriptsize
\centering
\setlength{\tabcolsep}{5.5pt} 
\caption{Comparisons of average MPJPEs across all actions in H3.6M. 
Red font indicates the better results in each method.
}
\vspace{-0.05in}
\label{tab:h3.6m}
\begin{tabular}{lccccccc}
\toprule
 \multirow{1}{*}{Mothod}  & 80ms & 160ms & 320ms & 400ms & 560ms & 1000ms &Average\\ \hline
 Traj-GCN & 12.19  &	24.87 &	50.76 &61.44 &	80.19 &	113.87 &57.22\\
 Traj-GCN-T & \textcolor{red}{11.31} &\textcolor{red}{24.10} &\textcolor{red}{49.95} &\textcolor{red}{60.72} & \textcolor{red}{78.44} &\textcolor{red}{113.00} & \textcolor{red}{56.25}\\ \hline
 SPGSN   & 10.74 &22.68 &47.46 &58.64 &79.88 &112.42 &55.30 \\
 SPGSN-T & \textcolor{red}{10.32} &\textcolor{red}{22.13} &\textcolor{red}{46.65} &\textcolor{red}{57.87} &\textcolor{red}{79.17}	&\textcolor{red}{112.08} &	\textcolor{red}{54.71}\\  \hline
 EqMotion  & 9.45 &21.01 &46.06 &\textcolor{red}{57.60} &75.98 &\textcolor{red}{109.75} &53.31\\ 
 EqMotion-T& \textcolor{red}{8.96 }&\textcolor{red}{20.50} &\textcolor{red}{45.93} &57.99 &\textcolor{red}{75.91} &109.76 &\textcolor{red}{53.01}\\ \hline
 STBMPT   & 10.73 &23.70 &49.72 &61.21 &82.26 &113.85 &56.91\\
 STBMPT-T &  \textcolor{red}{10.29} &\textcolor{red}{22.98} &\textcolor{red}{48.85} &\textcolor{red}{60.71} & \textcolor{red}{80.97} 	&\textcolor{red}{113.05} &\textcolor{red}{56.14}\\ \hline
 STBMPS   & 9.56 &21.80 &47.18 &58.64 &80.75 &113.98 &55.32\\
 STBMPS-T & \textcolor{red}{9.44 }&\textcolor{red}{21.69} &\textcolor{red}{47.09} &\textcolor{red}{58.56} &\textcolor{red}{78.36} &\textcolor{red}{111.83} &\textcolor{red}{54.49}\\  
\bottomrule
\end{tabular}
\vspace{-0.1in}
\end{table}

\begin{table}[!t]\scriptsize
\centering
\setlength{\tabcolsep}{5.5pt} 
\caption{Comparisons of average MPJPEs across all actions in the CMU-Mocap dataset.
}
\vspace{-0.05in}
\label{table:CMU}
\begin{tabular}{lccccccc}
\toprule
 \multirow{1}{*}{Mothod}  & 80ms & 160ms & 320ms & 400ms & 560ms & 1000ms &Average\\ \hline
Traj-GCN   & 11.34 &19.96 &37.64 &46.57 &62.15 &96.91 &45.76\\
Traj-GCN-T &  \textcolor{red}{10.80} &\textcolor{red}{19.13} &\textcolor{red}{36.60} &\textcolor{red}{45.37} &\textcolor{red}{60.36} &\textcolor{red}{93.93} &\textcolor{red}{44.37} \\ \hline
SPGSN      &  10.83 &19.72 &36.84 &44.94 &59.37 &88.00 &43.28\\
SPGSN-T    &  \textcolor{red}{9.92 }&\textcolor{red}{18.06} &\textcolor{red}{33.76} &\textcolor{red}{41.16} &\textcolor{red}{54.30} &\textcolor{red}{82.58} &\textcolor{red}{39.96}  \\ \hline
STBMPT   & 9.10 &16.47 &31.72 &39.49 &53.73 &79.48 &38.33\\
STBMPT-T &  \textcolor{red}{8.72} &\textcolor{red}{16.12} &\textcolor{red}{31.64} &\textcolor{red}{39.36} &\textcolor{red}{53.49} &\textcolor{red}{79.40} &\textcolor{red}{38.12} \\ \hline 
STBMPS   & 8.76 &16.54 &33.77 &42.43 &57.99 &86.70 &41.03\\
STBMPS-T & \textcolor{red}{8.47} &\textcolor{red}{16.02} &\textcolor{red}{32.52} &\textcolor{red}{40.65} &\textcolor{red}{55.32} &\textcolor{red}{83.85} &\textcolor{red}{39.47} \\
\bottomrule
\end{tabular}
\vspace{-0.15in}
\end{table}

\subsection{Datasets and baselines}\label{sec:data}
We evaluate our method on the Human3.6m (H3.6M)~\cite{ionescu2013human3} and CMU Motion Capture (CMU-Mocap)~\cite{dang2021msr,li2022skeleton} datasets. 
For comparison, we use several state-of-the-art and open-source baselines, including Traj-GCN~\cite{mao2019learning}, SPGSN~\cite{li2022skeleton}, Eqmotion~\cite{xu2023eqmotion}, STBMP~\cite{wang2023spatio}.  Baselines integrated with our approach are denoted with the suffix "-T" ($e.g.$, SPGSN-T).
Mean Per Joint Position Error (MPJPE) is reported to evaluate the performance, the lower indicates better performance.

\subsection{Experiment Results}

\noindent\textbf{Motion Prediction.}  
To validate our method, we carry out five experiments for each method on all datasets, with results reported as average scores. STBMPT and STBMPS represent the temporal and spatial branches of STBMP without incremental information. 
Tab.~\ref{tab:h3.6m} and Tab.~\ref{table:CMU} show prediction performance at various time steps. Models incorporating our method generally outperform the original models, demonstrating its effectiveness in refining motion predictions and advancing human motion prediction tasks.
Specifically, our method improves average performance by 1.84\% for Traj-GCN, 1.08\% for SPGSN, 0.08\% for EqMotion, 1.35\% for STBMPT and 1.48\% for STBMPS in H3.6M, and by 3.05\% for Traj-GCN, 6.75\% for SPGSN,  0.54\% for STBMPT and 3.80\% for STBMPS in CMU-Mocap. 
These results confirm that our method contributes to improvements in most time intervals, with a consistent improvement in the average performance. 

\begin{figure}[!t]
\begin{center}
\vspace{-0.1in}
\subfigure[EqMotion.]{
\includegraphics[width=0.22\columnwidth]{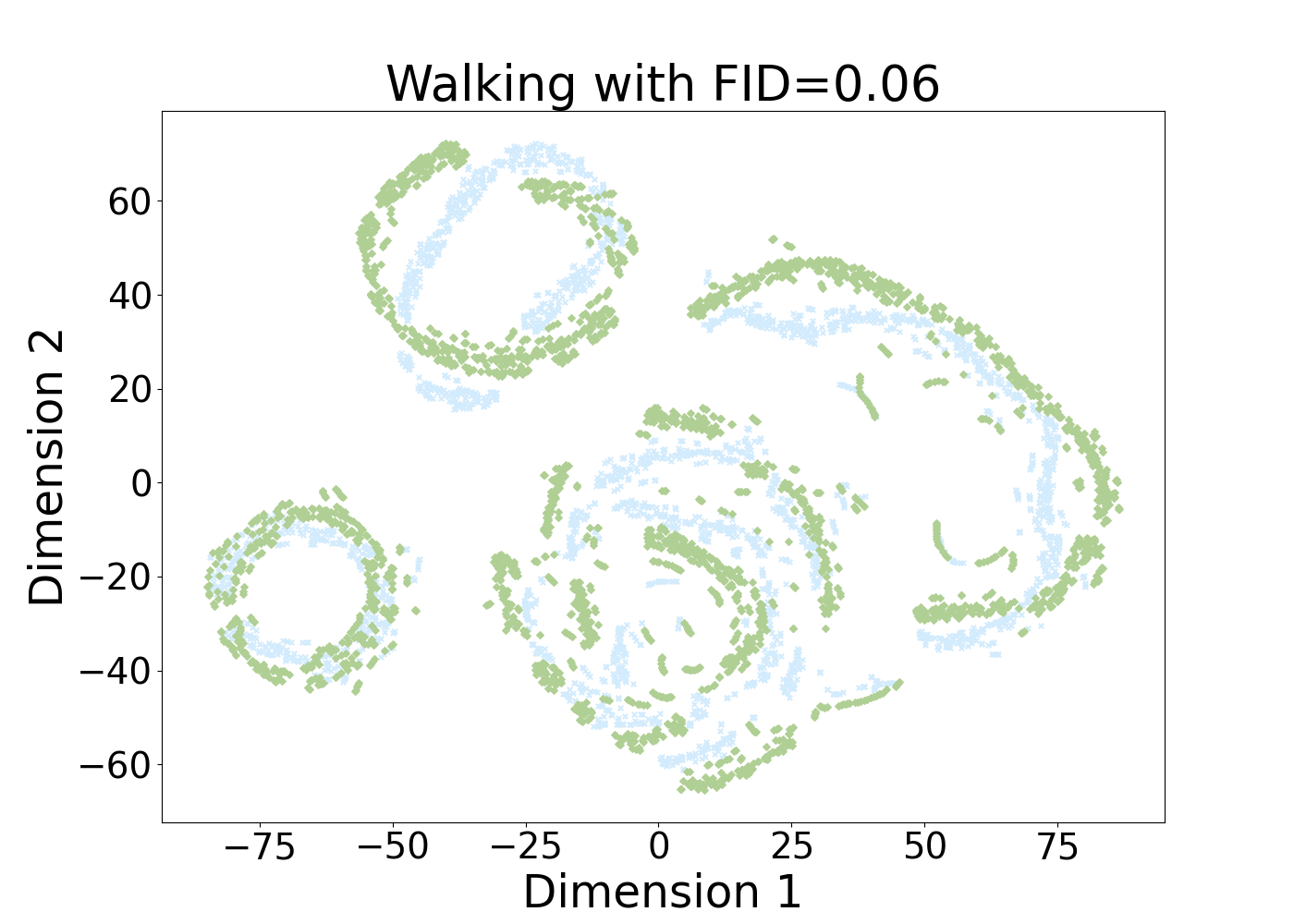}
\includegraphics[width=0.22\columnwidth]{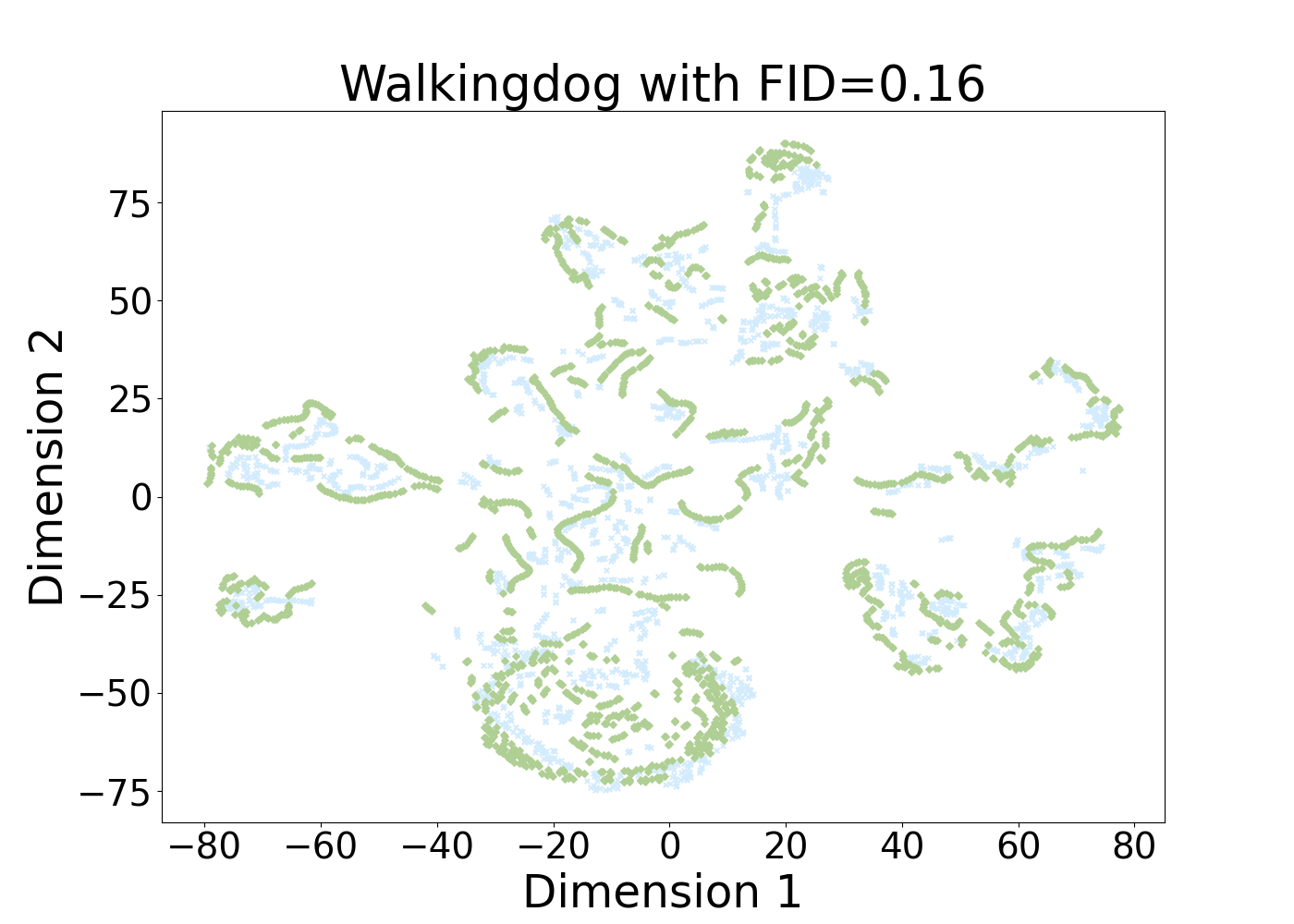}
}
\subfigure[EqMotion with our method.]{
\includegraphics[width=0.22\columnwidth]{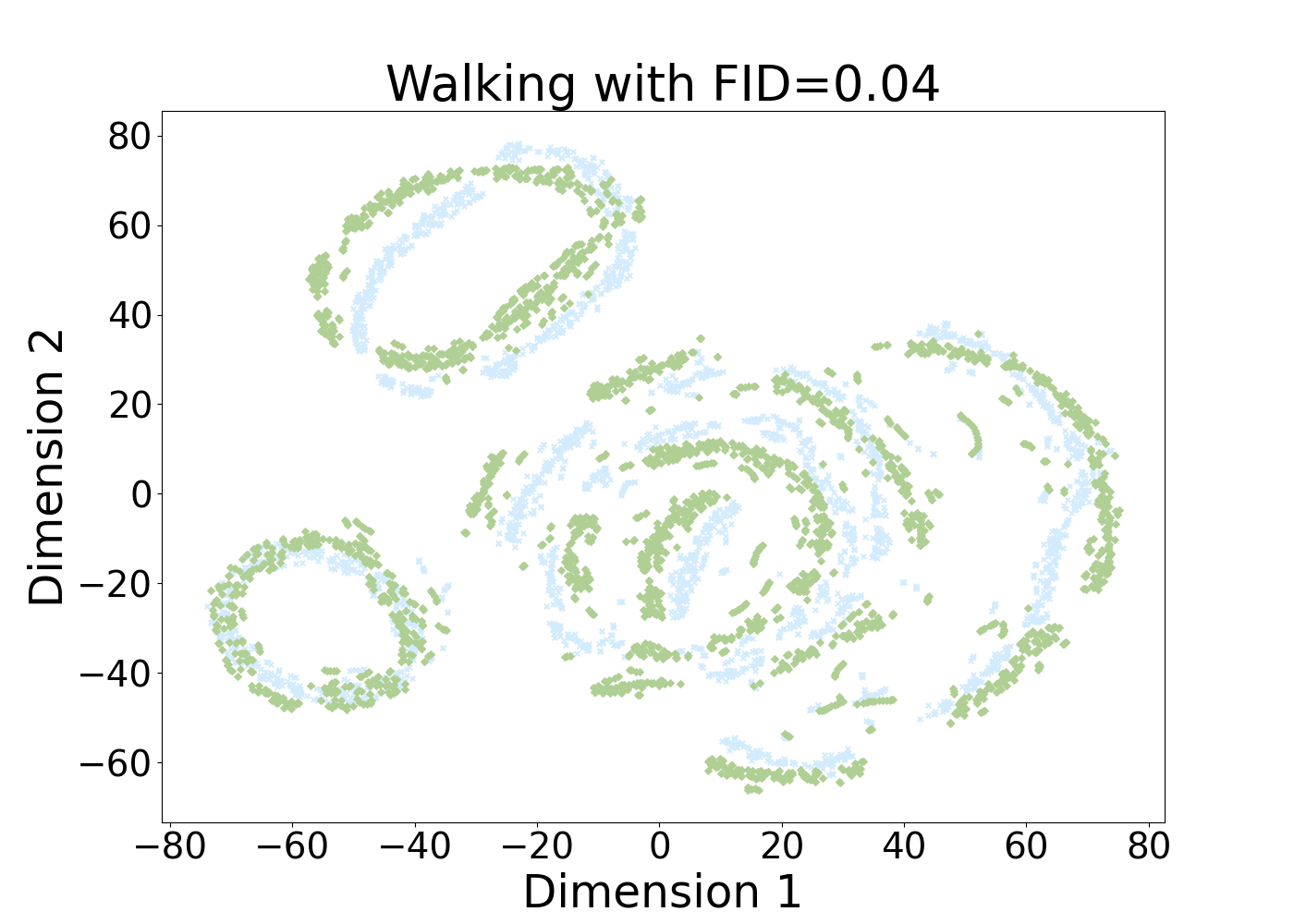}
\includegraphics[width=0.22\columnwidth]{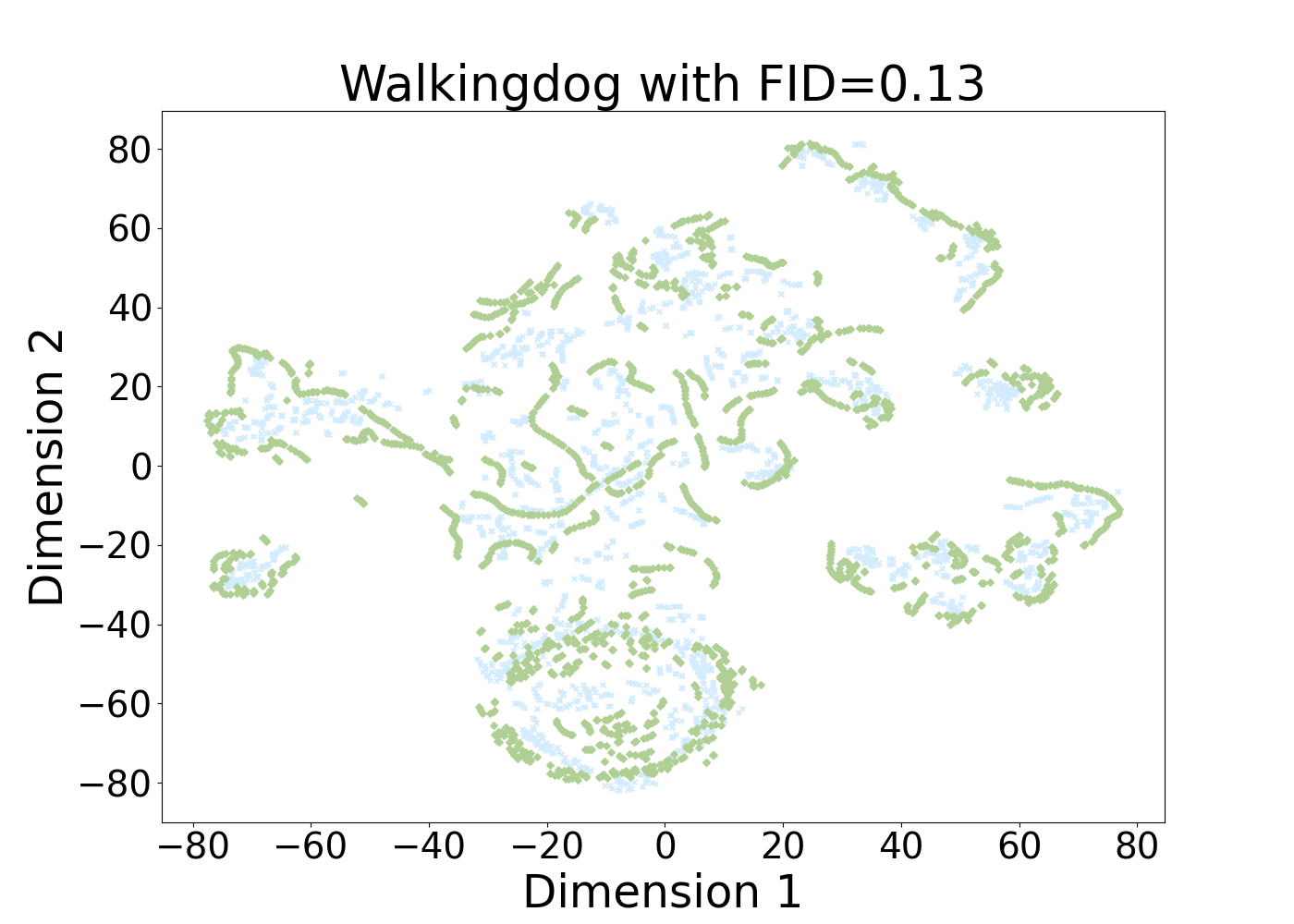}
}
\vspace{-0.1in}
\caption{T-SNE visualization of human motion. Green represents the ground truth motion features, and blue depicts the motion features predicted by the model. }
\label{fig:feature_compare}
\vspace{-0.2in}
\end{center}
\end{figure}

\begin{figure}[!t]
\begin{center}
\subfigure[``Purchases'' in EqMotion.]{
\includegraphics[width=0.45\columnwidth]{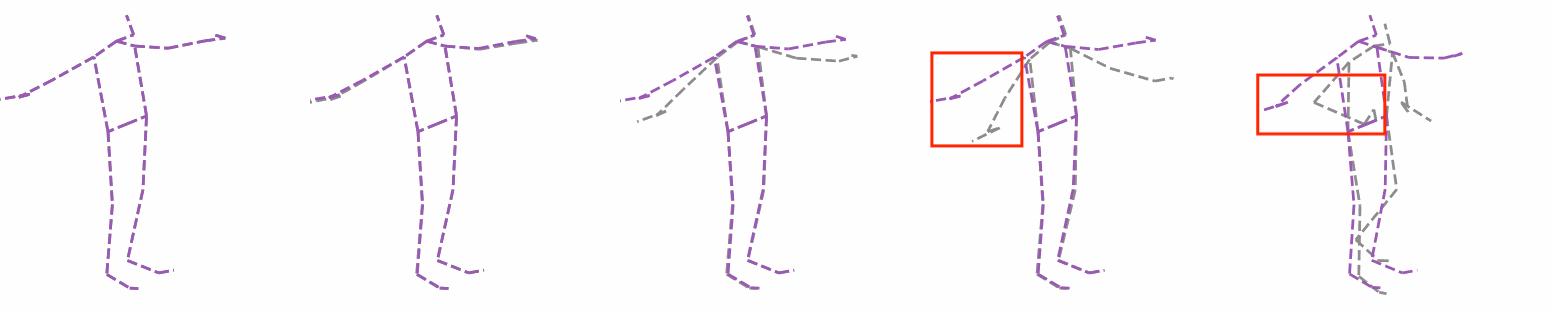}
}
\subfigure[``Purchases'' in EqMotion-T.]{
\includegraphics[width=0.45\columnwidth]{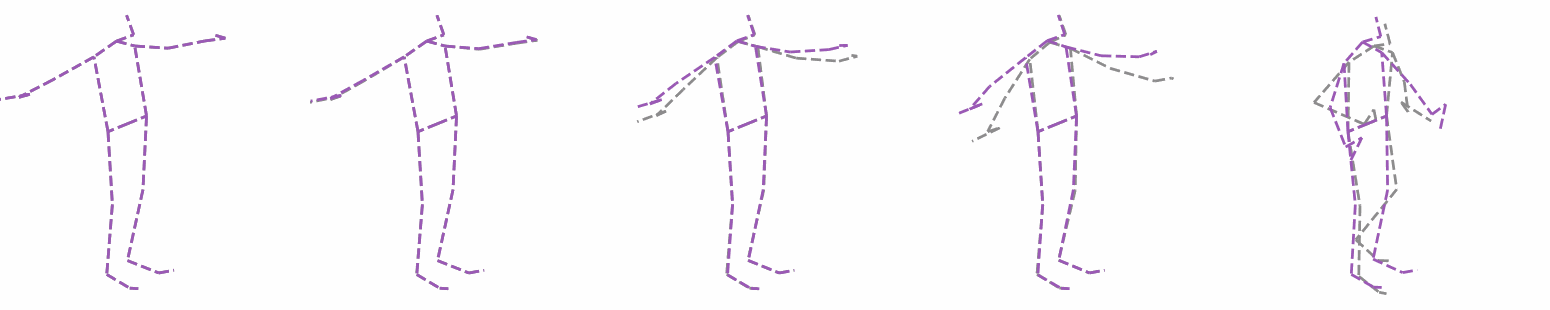}
}
\subfigure[``Walkingdog'' in SPGSN.]{
\includegraphics[width=0.45\columnwidth]{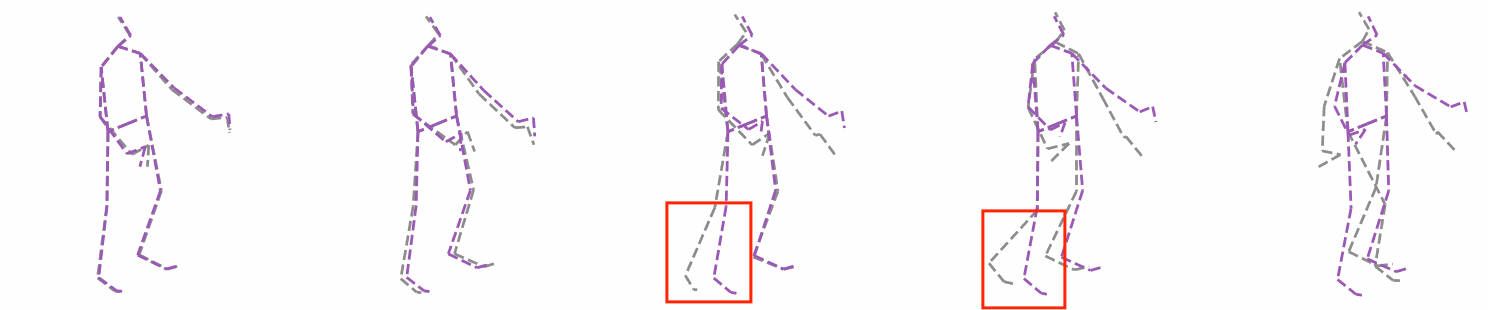}
}
\subfigure[``Walkingdog'' in SPGSN-T.]{
\includegraphics[width=0.45\columnwidth]{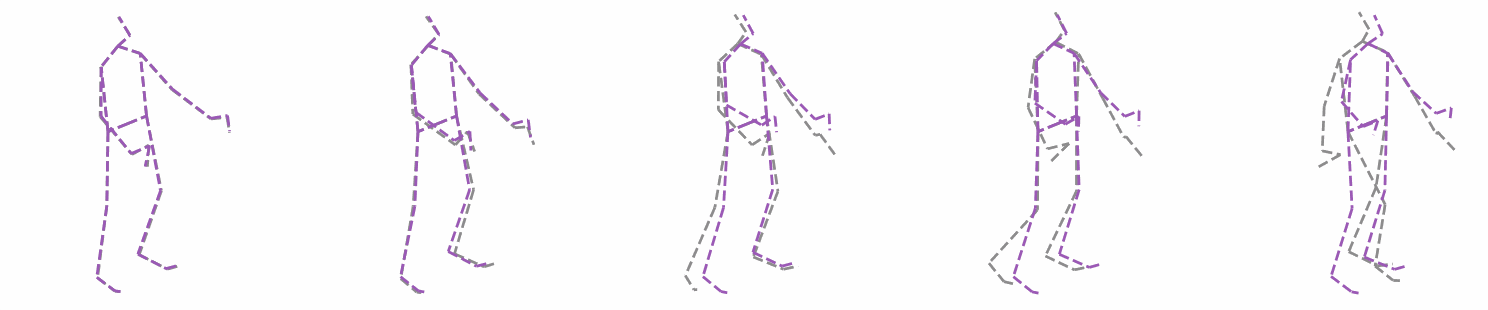}
}
\vspace{-0.1in}
\caption{Prediction samples on H3.6M for 80, 160, 320, 400 and 1000 ms. The purple dotted lines indicates the predictions and the grey lines indicate the ground truth actions.}
\label{appendix:viz}
\vspace{-0.15in}
\end{center}
\end{figure}

\noindent\textbf{Visualization Results.}
To illustrate the effectiveness of our method, we present visual results in  Fig~\ref{fig:feature_compare}. Specifically, we use T-SNE to visualize the motion features in EqMotion. This visualization confirms that our method brings the predicted action features closer to the ground truth features. Besides, we use the Frechet Inception Distance (FID) to quantify the dissimilarity between the predicted and ground truth motion feature distributions, where a smaller FID value signifies a closer match in distribution. The results demonstrate a reduction in FID values for our predictions, reflecting the improved performance of our approach. 
To further qualitatively evaluate the predictions, we depict the prediction of ``Purchases'', and ``Walkingdog'' in Fig{~\ref{appendix:viz}}, where the original model tends to exhibit larger prediction errors in the arms and legs and may output the ``mean pose". 

\begin{table}[!t]\scriptsize
\setlength{\tabcolsep}{7.5pt}
\centering
\caption{Ablation of our methods: Average MPJPE for short-term prediction (80/160/320/400ms) in H3.6M.}
\vspace{-0.05in}
\label{tab:ab_architecture}
\begin{tabular}{ccc|cccc}
\toprule
$\mathcal{L}_f$ &$\mathcal{L}_r$ &$TDD$ & Traj-GCN &SPGSN &EqMotion  & Average\\ \hline 
$\checkmark$& & & 37.31 &34.88 &33.53  &35.24 \\
$\checkmark$ & & $\checkmark$ & 36.93& 34.67 &33.52 & 35.04\\
$\checkmark$ &$\checkmark$ & & 36.29 & 34.49 & \textcolor{red}{33.29} &34.69\\ 
&$\checkmark$ &$\checkmark$ &  41.23 &37.91 &37.13 &38.76 \\ 
$\checkmark$ & $\checkmark$ &  $\checkmark$ & \textcolor{red}{36.52} &\textcolor{red}{34.24} & 33.34  & \textcolor{red}{34.70}\\ 
\bottomrule  
\end{tabular}
\vspace{-0.1in}
\end{table}

\noindent \textbf{Ablation Study}
To better understand our approach, we analyze the individual components and configurations of our proposed method. Tab.~\ref{tab:ab_architecture} shows comprehensive ablation experiments on various variants of the entire model. We can observe that i) the incorporation of $TDD$ and $IP$ ($\mathcal{L}_r$) individually leads to a significant improvement in the model performance, underscoring the efficacy of these components; ii) the combination of $TDD$ and $IP$ further improves the model’s performance, confirming the collaborative effects of $TDD$ and $IP$. These results underscore the importance of improving motion prediction by decoupling the decoding process for reconstruction and prediction and considering the bidirectional temporal correlation of motion.
\section{Conclusion}
\label{sec:conclusion}
In this study, we propose $TD^2IP$, a novel extension to the mainstream Encoder-Decoder framework for human motion prediction. 
Going beyond the previous convention that directly incorporates the reconstruction task into the decoder, our approach decouples the reconstruction and prediction decoding process, effectively alleviating the interference and conflicts between both tasks. 
Additionally, we propose an innovative auxiliary task called inverse processing, enabling each decoder to access complete motion information, fostering a more comprehensive understanding of human motion behaviors and enhancing the bidirectional correlation between historical and future behaviors.
Our approach maintains simplicity and effectiveness, ensuring practicality and efficiency, and seamlessly integrates with various prediction methods. Extensive experiments on human motion prediction benchmarks validate the effectiveness and superiority of our method, positioning it as a valuable alternative in the field. 

\section*{Acknowledgment}
\addcontentsline{toc}{section}{Acknowledgment}
This work was supported in part by the National Natural Science Foundation of China No. 62376277 and Public Computing Cloud, Renmin University of China.

\bibliographystyle{IEEEtran}
\bibliography{IEEEabrv}

\end{document}